\pdfoutput=1
\documentclass[11pt,a4paper]{article}
\usepackage[hyperref]{acl2020}
\usepackage{times}
\usepackage{latexsym}

\usepackage{pifont}
\newcommand{\cmark}{\ding{51}}

\usepackage{microtype}

\usepackage[hang,flushmargin]{footmisc}  
\usepackage{graphicx}
\usepackage{listings}
\usepackage{fancyvrb}
\usepackage{color}
\aclfinalcopy 

\pdfoutput=1

\newcommand\LN{\linebreak\noindent}
\newcommand\NATEX{\textsc{Natex}}

\title{Emora STDM: A Versatile Framework for\\ Innovative Dialogue System Development}

\author{James D. Finch \\
  Department of Computer Science \\
  Emory University \\
  Atlanta, GA, USA \\
  \texttt{jdfinch@emory.edu} \\\And
  Jinho D. Choi \\
  Department of Computer Science \\
  Emory University \\
  Atlanta, GA, USA \\
  \texttt{jinho.choi@emory.edu} \\}

\date{}

\begin{document}
\maketitle

\begin{abstract}
This demo paper presents Emora STDM (State Transition Dialogue Manager), a dialogue system development framework that provides novel workflows for rapid prototyping of chat-based dialogue managers as well as collaborative development of complex interactions.
Our framework caters to a wide range of expertise levels by supporting interoperability between two popular approaches, state machine and information state, to dialogue management. 
Our Natural Language Expression package allows seamless integration of pattern matching, custom NLP modules, and database querying, that makes the workflows much more efficient.
As a user study, we adopt this framework to an interdisciplinary undergraduate course where students with both technical and non-technical backgrounds are able to develop creative dialogue managers in a short period of time. 
\end{abstract}

\section{Introduction}

Constructing a functional end-to-end dialogue system is typically an extensive development process. 
Depending on the goals, such development often involves defining models for natural language understanding and generation (Section~\ref{sec:natex}), and also creating dialogue management logic to control conversation flow. 
Training a deep learning-based end-to-end model is a cost-effective way to develop a dialogue agent when the goal is a system conforming to behaviors present in training data; however, substantial development effort must be spent as the developer demands broaden to incorporate features that are not well-represented in available data.

We present Emora STDM (State Transition Dialogue Manager), henceforth E-STDM, a dialogue system development framework that offers a high degree of customizability to experts while preserving a workflow intuitive to non-experts.
E-STDM caters to a wide range of technical backgrounds by supporting the interoperability between two popular dialogue management approaches, state machine and information state (Section~\ref{sec:dialogue-management}).
Our framework makes it easy for not only rapid prototyping of open-domain and task-oriented dialogue systems, but also efficient development of complex dialogue managers that tightly integrate pattern matching, NLP models, and custom logic such as database queries. (Section~\ref{sec:user-study}).

\begin{table*}[htbp!]
\centering\small
\begin{tabular}{c|c|c|c||c|c|c|c|c|c|c|c}
	\bf ID & \bf Framework & \bf Type & \bf License  & \bf SM & \bf IS & \bf PM & \bf IC & \bf EF & \bf ON & \bf ET & \bf CM \\ \hline\hline
	  1    &  Emora STDM   & Library  &  Apache 2.0  & \cmark & \cmark & \cmark & \cmark & \cmark & \cmark & \cmark & \cmark \\ \hline
	  2    &     AIML      & Language &   GNU 3.0    &        &        & \cmark &        &        &        &        &        \\ \hline
	  3    &  RiveScript   & Language &     MIT      &        &        & \cmark &        & \cmark &        &        & \cmark \\ \hline
	  4    &  ChatScript   & Language &     MIT      & \cmark &        & \cmark &        & \cmark & \cmark &        &        \\ \hline
	  5    &     botml     & Language &     MIT      & \cmark &        & \cmark &        & \cmark &        &        &        \\ \hline
	  6    &   OpenDial    &   Tool   &     MIT      &        & \cmark & \cmark &        & \cmark &        &        &        \\ \hline
	  7    &    PyDial     &   Tool   &  Apache 2.0  &        & \cmark &        &        & \cmark & \cmark &        & \cmark \\ \hline
	  8    &     VOnDA     &   Tool   & CC BY-NC 4.0 &        & \cmark & \cmark &        & \cmark & \cmark &        &        \\ \hline
	  9    &   Botpress    &   Tool   &  Commercial  & \cmark &        &        & \cmark & \cmark &        & \cmark &        \\ \hline
	  10   &     RASA      &   Tool   &  Commercial  & \cmark &        &        & \cmark & \cmark &        & \cmark &        \\ \hline
	  11   &  DialogFlow   &   API    &  Commercial  & \cmark &        &        & \cmark &        &        & \cmark &
\end{tabular}
\caption{Comparison of features supported by various dialogue system development frameworks. SM: state machine, IS: information state, PM: pattern matching for natural language, IC: developer-trained intent classification, EF: external function calls, ON: ontology, ET: error tracking, CM: combine independent dialogue systems.}
\label{tab:comparison}
\vspace{-2ex}
\end{table*}

\vspace{-0.5ex}
\section{Related Work}
\vspace{-0.5ex}

A variety of dialogue development frameworks have emerged to expedite the process of dialogue system creation.
These frameworks cater to various use cases and levels of developer expertise. 
Popular commercial-oriented frameworks are primarily intended for non-experts and have workflows supporting rapid prototyping \cite{Bocklisch_Faulkner_Pawlowski_Nichol_2017}.
They often allow developers to customize natural language understanding (NLU) modules and perform dialogue management using state machines. 

Some frameworks require more expertise, but offer better developer control, by following the information state formulation of dialogue management \cite{Ultes_Rojas_Barahona_Su_Vandyke2017,Jang_Lee_Park_Lee_Lison_Kim_2019,Kiefer_Welker_Biwer_2019}. 
According to this formulation, dialogues are driven by iterative application of logical implication rules \cite{Larsson_Traum_2000}. 
This design provides support for complex interactions, but sacrifices the intuitiveness and development speed of dialogue management based on state machines.


Other frameworks (e.g., ChatScript, botml) rely on custom programming languages to design conversation flow.
The custom syntax they specify is based on pattern matching. 
Although requiring expertise, rapid prototyping in these frameworks is possible with a high degree of developer's control. 
However, dialogue management focuses primarily on shallow pattern-response pairs, making complex interactions more difficult to model.

\noindent Table~\ref{tab:comparison} shows a comparison of E-STDM to existing frameworks.
E-STDM is most similar to PyOpenDial and botml, which support pattern matching for NLU and tight integration of external function calls. 
Unlike any existing framework, however, E-STDM explicitly supports both state machine and information state paradigms for dialogue management, and also provides NLU that seamlessly integrates pattern matching and custom modules.\footnote{Emora STDM is available as an open source project at\\\url{github.com/emora-chat/emora\_stdm}.}

\section{\NATEX: Natural Language Expression}
\label{sec:natex}

To address the challenge of understanding user input in natural language, we introduce the NATural language EXpression, \NATEX, that defines a comprehensive grammar to match patterns in user input by dynamically compiling to regular expressions. 
This dynamic compilation enables abstracting away unnecessary verbosity of regular expression syntax, and provides a mechanism to embed function calls to arbitrary Python code.
We highlight the following key features of \NATEX.


\paragraph{String Matching} 
It offers an elegant syntax for string matching.
The following \NATEX\ matches user input such as `\textit{I watched avengers}' or `\textit{I saw Star Wars}' and returns the variable \texttt{\$MOVIE} with the values `\textit{Avengers}' and `\textit{Star wars}', respectively:

\begin{center}
\small\begin{BVerbatim}
[I {watched, saw}
 $MOVIE={Avengers, Star Wars}]
\end{BVerbatim}
\end{center}

\noindent The direct translation of this \NATEX\ to a regular expression would be as follows:

\begin{center}
\small\begin{BVerbatim}
.*?\bI\b
.*?(?:\b(?:watched)\b|\b(?:saw)\b)
.*?(?P<MOVIE>(?:\b(?:avengers)\b|
                \b(?:star wars)\b)).*?
\end{BVerbatim}
\end{center}

\noindent As shown, \NATEX\ is much more succinct and interpretable than its counterpart regular expression. 


\paragraph{Function Call}
It supports external function calls.
The following \NATEX\ makes a call to the function \texttt{\#MDB} in Python that returns a set of movie titles: 

\begin{center}
\small\begin{BVerbatim}
[I {watched, saw} $MOVIE=#MDB()]
\end{BVerbatim}
\end{center}

\noindent This function can be implemented in various ways (e.g., database querying, named entity recognition), and its \NATEX\ call matches substrings in the user input to any element of the returned set.
Note that not all elements are compiled into the resulting regular expression; only ones that are matched to the user input are compiled so the regular expression can be processed as efficiently as possible. 




\paragraph{Ontology} 
It supports ontology editing and querying as the built-in \NATEX\ function called \texttt{\#ONT}.
An ontology can be easily built and loaded in JSON. 
\texttt{\#ONT(movie)} in the example below searches for the \texttt{movie} node in a customizable ontology represented by a directed acyclic graph and returns a set of movie titles from the subgraph of \texttt{movie}:

\begin{center}
\small\begin{BVerbatim}
[I {watched, saw} $MOVIE=#ONT(movie)]
\end{BVerbatim}
\end{center}



\paragraph{Response Generation}

It can be used to generate system responses by randomly selecting one production of each disjunction (surrounded by \texttt{\{\}}) in a top-down fashion. 
The following \NATEX\ can generate ``\textit{I watched lots of action movies lately}'' or ``\textit{I watched lots of drama movies recently}'', and assign the values of `\textit{action}' and `\textit{drama}' to the variable \texttt{\$GENRE} respectively:

\begin{center}
\small\begin{BVerbatim}
I watched lots of $GENRE={action, 
horror, drama} movies {recently, lately}
\end{BVerbatim}
\end{center}



\begin{figure*}
    \centering
    \includegraphics[width=\textwidth]{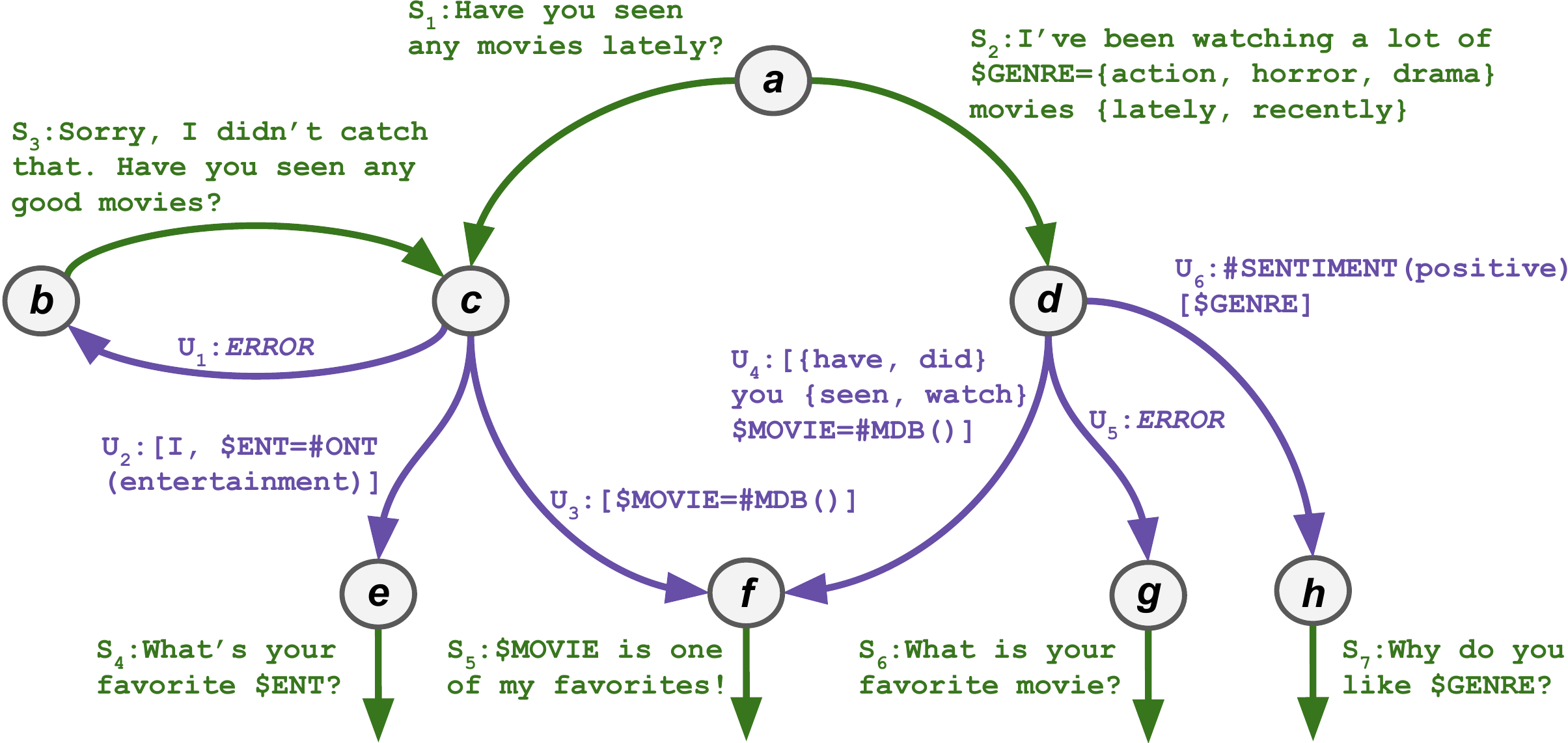}
    \caption{A dialogue graph using a state machine approach with \NATEX\ to dialogue management.}
    \label{fig:corr}
\end{figure*}

\paragraph{Error Checking}

Our \NATEX\ compiler uses the Lark parser to automatically detect syntax errors.\footnote{\url{https://github.com/lark-parser/lark}}
Additionally, several types of error checking are performed before runtime such as:

\begin{itemize}
\setlength\itemsep{0em}
    \item Call to a non-existing function.
    \item Exceptions raised by any function.
    \item Function returning mismatched type.
    \item Reference to a non-existing variable.
\end{itemize}

\section{Dialogue Management}
\label{sec:dialogue-management}

\subsection{Dialogue State Machine}
\label{ssec:state-machine}

The primary component responsible for dialogue management within E-STDM is a state machine. 
In our framework, state transitions alternate between the user and the system to track turn taking, and are defined by \NATEX\ (Figure~\ref{fig:corr}). 
Any transition performed with a variable-capturing \NATEX\ will store\LN a variable-value pair in a dedicated table in memory, which persists globally for future reference.

User turns are modeled by transitions according to which \NATEX\ matches the user input. 
To resolve\LN cases where multiple \NATEX\ yield matches, transitions can be defined with priority values. 
Similarly, developers can specify a catch-all ``error transition'' (\texttt{ERROR} in Figure~\ref{fig:corr}) to handle cases where no transition's \NATEX\ returns a match. 
The user input resulting in an error transition is automatically logged to improve the ultimate design of the state machine. 

System turns are modeled by randomly selecting\LN an outgoing system transition. 
Random selection promotes uniqueness among dialogue pathways, but can be restricted by specifying explicit priority values. 
To avoid redundancy when returning to a previously visited state, E-STDM prefers system transitions that have not been taken recently.

\noindent The simplicity of this dialogue management formulation allows for rapid development of contextually aware interactions. 
The following demonstrates the streamlined JSON syntax for specifying transitions $S_{1}, S_{3}, U_{1}, U_2,$ and $U_3$ in Figure~\ref{fig:corr}. 

\begin{center}
\small\begin{BVerbatim}
{
  "Have you seen any movies lately?": {
    "state": "c",
    "[I, $ENT=#ONT(entertainment)]": {
      "What's your favorite $ENT?": {..}
    },
    "[$MOVIE=#MDB()]": {
        "$MOVIE is one of my ...": {..}
    }
    "error": {
        "Sorry, I didn't catch ...": "c"
    } } }
\end{BVerbatim}
\end{center}

\subsection{Information State Update Rules}
\label{ssec:information-state}

Despite its simplicity, state machine-based dialogue management often produces sparsely connected state graphs that are overly rigid for complex interactions \cite{Larsson_Traum_2000}. E-STDM thus allows developers to specify information state update rules to take advantage of the power of information state-based dialogue management.

Information state update rules contain two parts, a precondition and a postcondition. 
Each user turn before E-STDM's state machine takes a transition, the entire set of update rules is iteratively evaluated with the user input until either a candidate system response is generated or no rule's precondition is satisfied.
In the following example, satisfying precondition {\small\texttt{[I have \$USER\_PET=\#PET()]}} triggers postcondition {\small\texttt{\#ASSIGN(\$USER\_LIKE=\$USER\_PET)}} to assign {\small\texttt{\$USER\_PET}} to {\small\texttt{\$USER\_LIKE}}, allowing rule {\small\texttt{\#IF(..)} \texttt{I like \$USER\_LIKE} \texttt{..}} to trigger in turn:

\begin{center}
\small\begin{BVerbatim}
{
  "[I have $USER_PET=#PET()]"
  : "#ASSIGN($USER_LIKE=$USER_PET)",
  "[$USER_FAVOR=#PET() is my favorite]"
  : "#ASSIGN($USER_LIKE=$USER_FAVOR)",
  "#IF($USER_LIKE != None)"
  : "I like $USER_LIKE too! (0.5)"
}
\end{BVerbatim}
\end{center}

\noindent When a precondition is satisfied, the postcondition is applied through the language generation (Sec.~\ref{sec:natex}). 
If a real-number priority is provided in parentheses at the end of any \NATEX, the generated string becomes a candidate system response. A priority value higher than any outgoing system transition in the dialogue state machine results in the candidate becoming the chosen one; thus, no dialogue state machine transition is taken. 
Often however, a developer can choose to omit the priority value to use \NATEX\ purely as a state updating mechanism.

This formalism allows flexible interoperability between state machine-based and information state-based dialogue management. 
Given E-STDM, developers have the latitude to develop a system entirely within one of the two approaches, although we believe a mixed approach lends the best balance of development speed and dialogue sophistication.

\subsection{Combining Dialogue Modules}

E-STDM has explicit support for a team-oriented workflow, where independent dialogue modules can be easily combined into one composite system.\LN
Combining multiple modules requires specification of a unique namespace per module to enforce encapsulation of both errors and identifiers. 
The following is an example Python script combining dialogue systems \texttt{df1} and \texttt{df2} under namespaces \texttt{DF1} and \texttt{DF2}, respectively: 

\begin{center}
\small\begin{BVerbatim}
df1 = DialogueFlow('start_1')
df1.add_transitions('df1.json')
df2 = DialogueFlow('start_2')
df2.add_transitions('df2.json')

cdf = CompositeDialogueFlow('start')
cdf.add_module(df1, 'DF1')
cdf.add_module(df2, 'DF2')
cdf.add_user_transition(
  'DF1.stateX', 'DF2.stateY', 
  "[{film, movie}]")
\end{BVerbatim}
\end{center}

\noindent Moreover, inter-component transitions can be made between any two dialogue states to seamlessly combine modules together and allow smooth topic transitions for better user experience.

\pagebreak
\section{Educational Use of Emora STDM}
\label{sec:user-study}

As an application case study, we present the use of\LN E-STDM in an educational setting. 
E-STDM is deployed in an interdisciplinary undergraduate course called \textit{Computational Linguistics},\footnote{\url{github.com/emory-courses/cs329}} where dialogue system development within E-STDM is a part of the requirements. 
Students in this course come with varying levels of programming ability; many with little to no imperative programming experience. 

Students are tasked with the development of chat-based dialogue systems that can engage a user in 10+ turn conversations. 
At the time of writing, students have completed two assignments involving dialogue system creation. 
Students are grouped in teams, with at least one student with prior coding experience per team. 
Teams are free to select a domain, such as video games, sports, or technology, and are given two weeks for development. 

We make the unmodified version of dialogue systems from these students publicly available.\footnote{\url{github.com/emora-chat/emora_stdm_zoo}}
The successful use of E-STDM by novice programmers demonstrates the utility of this framework, in terms of its usability and potential as an educational tool.

\section*{Acknowledgments}

We gratefully acknowledge Sarah E. Finch for her support in developing E-STDM as well as assessing the course assignments (Section~\ref{sec:user-study}).

\bibliography{acl2020}
\bibliographystyle{acl_natbib}


\end{document}